\documentclass[11pt]{article}

\usepackage[final]{acl}

\usepackage{times}
\usepackage{latexsym}

\usepackage[T1]{fontenc}

\usepackage[utf8]{inputenc}

\usepackage{microtype}

\usepackage{inconsolata}

\usepackage{graphicx}

\usepackage{amsmath}
\usepackage{amssymb}
\usepackage{booktabs}
\usepackage{siunitx}
\usepackage{multirow}

\usepackage{polyglossia}

\newfontfamily\devanagarifont{NotoSerifDevanagari}[
  Path      = ./fonts/,
  Extension = .ttf,
  UprightFont = *-Regular,
  BoldFont    = *-Bold,
  Script    = Devanagari,
]

\setmainlanguage{english}
\setotherlanguage{hindi}
\title{StalePO: Anchored Token-Level Preference Optimization using Legacy Post-Edits in Machine Translation}

\author{
  \textbf{Rohit Dhaipule},
  \textbf{Sukhdeep Singh Kharbanda},
  \textbf{Prasanth Bathala},
\\
  \textbf{Pradyumna Lanka},
  \textbf{Anubhav Shrimal}
\\
  Translation Services, Amazon
\\
  \small{
    \{rdhaipu, kharsukh, pbathala, pradyl, shrimaa\}@amazon.com
  }
}

\begin{document}
\maketitle
\begin{abstract}
Machine translation systems are periodically upgraded to stronger models, but the available preference signal is human post-edits of an older system's outputs, which the newer model may already surpass. Moreover, collecting fresh post-edits for every new model is prohibitively expensive. We call this the \textbf{Stale Preference} problem. Standard DPO can fail in this setting: it may increase the likelihood of inferior post-edits, erode the model's existing quality, and fail to provide the per-token control needed to correct localized errors. We introduce \textbf{StalePO}, an objective derived from three requirements this regime imposes. Likelihood movement must be downward on both responses, the policy must be anchored to its own base response, and the KL constraint must apply at the token level. These requirements are jointly necessary. In ablations, each mechanism in isolation leaves the model's performance indistinguishable from the base model, and only their combination converts stale feedback into gains. On English-to-Hindi and English-to-Turkish localization data, StalePO improves the fraction of segments passing all LLM-as-judge MQM quality checks by 14.9 and 4.6 percentage points, respectively, with gains concentrated on style and fluency. A human evaluation under the same framework confirms these gains on English-to-Hindi, raising the fraction of segments passing all seven human checks by 13.8 percentage points.
\end{abstract}

\section{Introduction}
\label{sec:introduction}

\begin{figure}[t]
\centering
\includegraphics[width=0.75\linewidth]{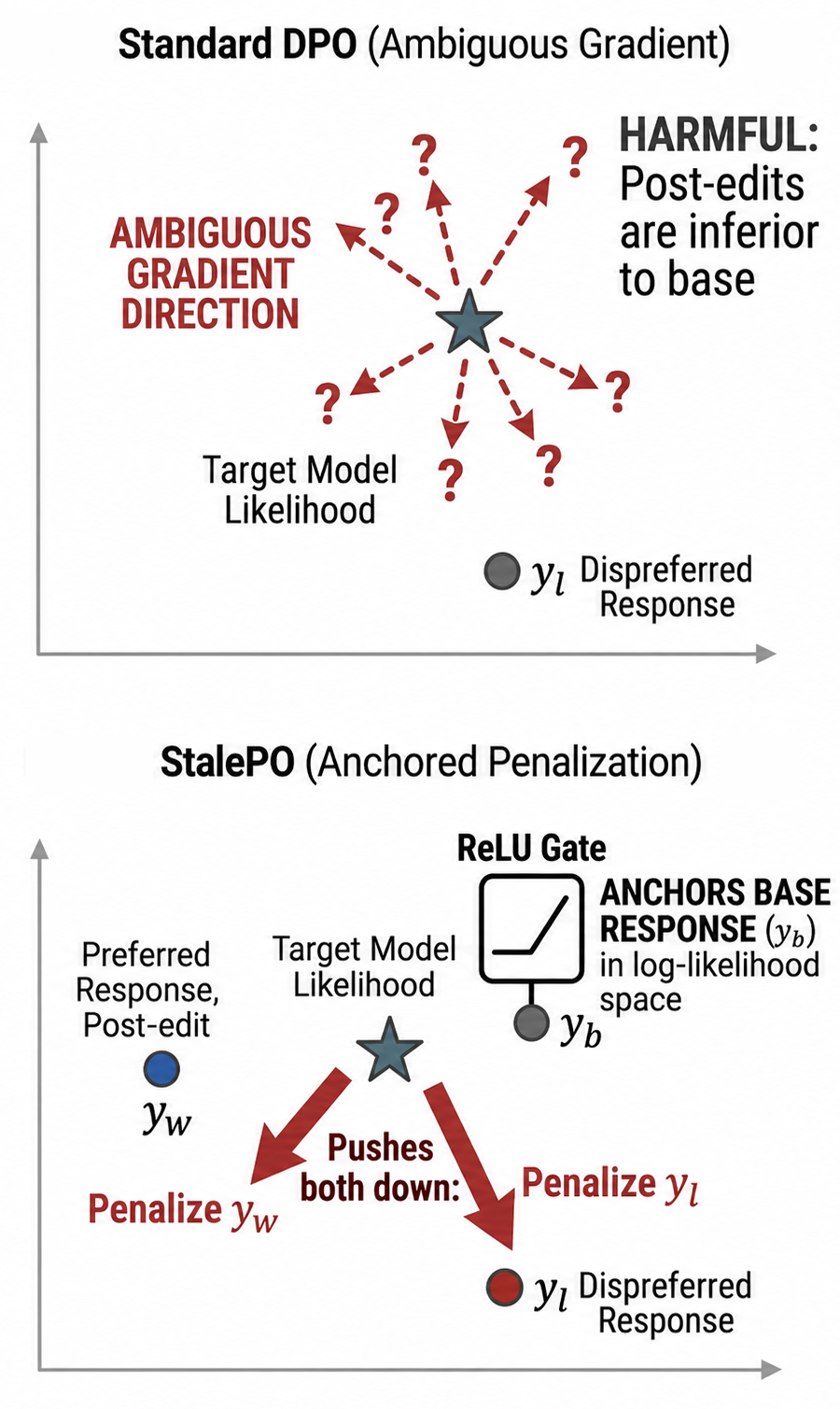}
\caption{
(Top) \textbf{DPO} increases the margin between the policy's log-likelihood of the preference pair ($y_w$, $y_l$) but does not constrain the direction of either change. (Bottom) \textbf{StalePO} is designed to encourage lower log-likelihoods for both $y_w$ and $y_l$ while anchoring the policy to its own base response $y_b$. The arrows illustrate this tendency rather than a guaranteed update direction.}
\label{fig:intro}
\end{figure}

Machine translation systems are periodically upgraded as stronger models become available. Further improving such a model requires human feedback on its own outputs, yet the available feedback consists of human post-edits (MTPE) collected on the outputs of the weaker systems it replaced \citep{berger2024postedits}. Post-editors correct existing output rather than retranslate from scratch, and collecting fresh post-edits for every model version is prohibitively expensive in pipelines that serve millions of segments each month. The preference data available for training is therefore biased toward earlier systems. Its preferred responses are typically inferior overall to what the current model already produces, yet they remain informative, because they correct genuine errors along dimensions on which the current model still underperforms, such as adherence to a target style guide. We refer to this as the \emph{stale preference} problem.

Aligning large language models with human preferences is now a standard stage in model development \citep{christiano2017deep, ouyang2022training, bai2022training}. Reinforcement Learning from Human Feedback (RLHF) introduced the core pipeline: train a reward model on pairwise human preferences, then optimize the policy with Proximal Policy Optimization \citep{schulman2017proximal}. Direct Preference Optimization (DPO; \citealp{rafailov2023direct}) reparameterized this objective as a contrastive loss over preference pairs, eliminating both explicit reward modeling and online sampling, and its simplicity has made it the dominant approach for offline alignment. However, like most offline preference methods, DPO assumes that the preferred output represents a quality target the model should approach.

Stale preferences violate this assumption, and we formalize the setting in Section~\ref{sec:stale_preference}. Applying DPO in this regime exposes three structural limitations. It constrains only the \emph{relative} likelihood change between the two responses in a preference pair, not the absolute direction of either \citep{doosterlinck2025anchored, pal2024smaug, tajwar2025preference}, so it may increase the likelihood of an inferior post-edit. Its KL penalty operates only over tokens present in the preference pair, providing no mechanism to preserve the model's behavior on its own superior generations \citep{lee2024bapo}. Its implicit reward aggregates all token positions into a single divergence term, sacrificing the per-position control needed to target localized style errors without disturbing the rest of the sequence \citep{zeng2024token}. Section~\ref{sec:limitations} examines these limitations in detail. Prior work addresses each in isolation, but no single method resolves all three, and none was designed for the stale preference regime, where all three arise together.

When these three limitations arise concurrently, DPO is unable to extract a meaningful preference signal from the preference pairs. As a result, despite deliberately curating the training data to target dimensions on which the base model underperforms, standard DPO produces negligible improvement over the base model. Avoiding this outcome requires a single objective that addresses all three limitations at once, which we introduce as StalePO (Stale Preference Optimization). Our contributions are as follows:
\begin{enumerate}
  \item We formalize the \emph{stale preference} problem in Section \ref{sec:stale_preference}, in which the preferred outputs are inferior overall to the model's own generations while still carrying useful signal on the dimensions where the model underperforms.
  \item We propose Stale Preference Optimization (StalePO), which incorporates three mechanisms: a tendency toward downward likelihood movement on both responses, base-response anchoring via a one-sided penalty, and per-token KL decomposition.
  \item We evaluate on English$\rightarrow$Hindi and English$\rightarrow$Turkish localization data, improving the all-pass rate by 14.9\,pp and 4.6\,pp respectively, with an evaluation by professional translators confirming a 13.8\,pp gain.
  \item We show that preference optimization on stale data is ineffective unless directional control, anchoring, and token-level constraints are combined. Each mechanism, considered in isolation, leaves the model's performance indistinguishable from the base model.
\end{enumerate}

\section{The Stale Preference Data Problem}
\label{sec:stale_preference}

Staleness in preference data largely stems from the economics of post-editing. A localization pipeline serving millions of segments per month cannot afford a fresh round of human post-editing each time the underlying model is replaced, so the preference data available for fine-tuning was almost always produced against an earlier system. Because post-editors edit rather than fully retranslate, their edits stay tied to the lexical and syntactic patterns of that earlier model. As a result, these post-edits are typically inferior to the current model's outputs on most quality dimensions, while retaining an advantage on the few dimensions on which the current model remains deficient. We formalize this as follows. Let $\mathcal{K}$ be a set of translation quality dimensions and $Q_k(x, y)$ the quality of translation $y$ on dimension $k$. A preference dataset $\mathcal{D} = \{(x, y_w, y_l)\}$ is \emph{stale} with respect to policy $\pi_\theta$ if, for a non-negligible fraction of examples, there exists a small subset $\mathcal{S} \subset \mathcal{K}$ with $|\mathcal{S}| \ll |\mathcal{K}|$ such that $Q_k(x, y_w) > Q_k(x, y_b)$ for all $k \in \mathcal{S}$ and $Q_k(x, y_w) < Q_k(x, y_b)$ for all $k \in \mathcal{K} \setminus \mathcal{S}$, where $y_b \sim \pi_\theta(\cdot|x)$. Exploiting this dual signal requires an objective that learns from the preference ordering without imitating either response, a requirement that existing methods do not address.

\section{Background and Related Work}
\label{sec:limitations}

\textbf{Direct Preference Optimization (DPO)} \citep{rafailov2023direct} optimizes the log-ratio between preferred and dispreferred outputs via the objective
\begin{equation}
\begin{aligned}
\mathcal{L}_{\text{DPO}}(x, y_w, y_l; \theta) 
&= -\log \sigma\!\left(
r_\theta(x,y_w)-r_\theta(x,y_l)
\right)
\end{aligned}
\end{equation}
where $\sigma$ denotes the logistic function and $r_\theta(x, y) = \beta \log \frac{\pi_\theta(y|x)}{\pi_{\text{ref}}(y|x)}$ is the implicit reward defined relative to a frozen reference policy $\pi_{\text{ref}}$. While DPO ensures the preferred output receives higher relative likelihood, it does not control the absolute direction of either update, so both likelihoods may rise, both may fall, or they may move in opposite directions. When the \textit{preferred} output is in fact worse than the base model's output, this lack of directional control admits updates that raise the likelihood of an inferior post-edit, so respecting the preference ordering does not imply that the resulting policy improves.

\textbf{Anchored Preference Optimization (APO-Down) }\citep{doosterlinck2025anchored} directly addresses this issue by biasing optimization toward lower likelihoods for both the preferred and dispreferred outputs while preserving preference separation:

\begin{equation}
\begin{aligned}
\mathcal{L}_{\text{down}}^{\text{APO}}(x, y_w, y_l; \theta)
&= \sigma(r_\theta(x, y_w)) \\
&\quad - \sigma(r_\theta(x, y_w) - r_\theta(x, y_l))
\end{aligned}
\end{equation}

This reduces the directional ambiguity present in DPO by discouraging increases in absolute likelihood for both elements of the pair, without guaranteeing the sign of every update. APO-Down has two limitations, however. First, without an explicit anchor to the model's own behavior, repeated application erodes baseline generation quality, since the model receives no signal to preserve its prior competence. Second, the objective operates at the sequence level. In machine translation, many violations are localized to a small number of tokens while the majority of the sequence stays the same. Sequence-level objectives provide no mechanism to target such positions selectively.

\paragraph{Other Related Work} No prior objective is designed for preference data whose preferred response is inferior to the policy's own output. Loss-function variants such as IPO \citep{azar2024general}, SLiC-HF \citep{zhao2023slic}, KTO \citep{ethayarajh2024kto}, NCA \citep{chen2024nca}, ORPO \citep{hong2024orpo}, and SimPO \citep{meng2024simpo} modify the training objective or remove the reference model entirely, but all continue to treat $y_w$ as a target to approach, which is precisely what fails when $y_w$ is worse than the policy on most dimensions. Online and iterative methods \citep{guo2024oaif, wu2025sppo, chen2024spin, calandriello2024online, liu2024rso} address a different form of staleness, namely drift between the training distribution and the current policy, by refreshing the data during training. This leaves the present problem untouched, since regenerating candidates does not improve post-edits that are inferior by construction, and it additionally requires online generation or multi-stage pipelines that are impractical in a high-volume setting. Token-level credit assignment methods \citep{cao2024sepo, deng2024sparsepo, zhou2024treg, zhang2024fpo} show that uniform token weighting is suboptimal, but they govern where updates are applied rather than whether the target should be imitated. Finally, within machine translation, CPO \citep{xu2024cpo} maximizes likelihood on winning outputs, an effective strategy when chosen translations are high-quality but directly harmful in our setting, while distributionally robust \citep{cheng2026drdpo} and length-disentangling \citep{park2024disentangling} formulations address failure modes unrelated to stale preferences.

\section{Stale Preference Optimization}
\label{sec:stalepo}

Stale Preference Optimization (StalePO) addresses all three limitations within a single objective, through a directional bias, base-response anchoring, and token-level regularization.

\begin{equation}
\mathcal{L}_{\text{StalePO}} = \mathcal{L}_{\text{pref}} + \lambda \cdot \mathcal{L}_{\text{Anchor}}
\end{equation}

where:
\begin{align}
    \begin{split}
        \mathcal{L}_{\text{pref}} &= \sigma\!\big(\tilde{r}_\theta(x, y_w)\big) \\
        &\quad - \sigma\!\big(\tilde{r}_\theta(x, y_w) - \tilde{r}_\theta(x, y_l)\big)
    \end{split} \\[1ex]
    \begin{split}
        \tilde{r}_\theta(x, y) &= \beta \cdot \bigg( \log\frac{\pi_\theta(y|x)}{\pi_{\text{ref}}(y|x)} \\
        &\quad + \alpha \cdot D_{\text{SeqKL}}(x, y; \pi_{\text{ref}} \| \pi_\theta) \bigg)
    \end{split} \\[1ex]
    \mathcal{L}_{\text{Anchor}} &= \max\left(0, \log \frac{\pi_{\text{ref}}(y_b|x)}{\pi_\theta(y_b|x)}\right)
\end{align}

Here $y_b$ denotes the model's own base response to source $x$ (sampled from the policy before training, see Section~\ref{sec:experiments}), and $D_{\text{SeqKL}}$ is the sequential forward KL divergence defined in Equation~\ref{eq:seqkl}, applied symmetrically to $y_w$ and $y_l$ with no stop-gradient on either side (at $\alpha{=}1$ this recovers TDPO1).

\paragraph{Directional Control} The preference term is designed to bias optimization toward lower likelihoods for both responses. Specifically, $\sigma(\tilde{r}_\theta(x, y_w))$ discourages the probability of the preferred response, while the difference term $-\sigma(\tilde{r}_\theta(x, y_w) - \tilde{r}_\theta(x, y_l))$ favors a larger penalty on the dispreferred one. This preserves their relative ordering without requiring an explicit increase in the preferred likelihood, thereby avoiding the directional ambiguity present in DPO.

\paragraph{Base-Response Preservation} The anchor term $\mathcal{L}_{\text{Anchor}}$ imposes a one-sided constraint on $y_b$. It activates only when the policy assigns lower likelihood to $y_b$ than the reference model, producing a gradient that pushes the likelihood back toward the reference, and contributes nothing otherwise. This ReLU-gated structure acts as a selective safeguard, preventing degradation of baseline behavior without interfering with preference optimization \citep{lee2024bapo}.

\paragraph{Token-Level Regularization} When the preference term lowers the likelihood of either response, neither component above controls where the displaced probability mass goes. Both are computed from sequence-level log-ratios, which track only the total log-probability of a complete response. The policy can therefore satisfy them while concentrating each next-token distribution on a few dominant tokens. Repeated over training, such updates induce token-level mode collapse, eroding the generation diversity of the base model. To counter this, we introduce a sequential forward KL divergence:

\begin{equation}
\label{eq:seqkl}
\begin{split}
D_{\text{SeqKL}}&(x, y; \pi_{\text{ref}} \| \pi_\theta) = \\
&\sum_{t=1}^{T} D_{\text{KL}}\big(\pi_{\text{ref}}(\cdot|[x, y^{<t}]) \| \pi_\theta(\cdot|[x, y^{<t}])\big)
\end{split}
\end{equation}

Following the TDPO1 formulation of \citet{zeng2024token}, this shifts standard DPO from aggregated token log-ratios to per-token constraints, ensuring that local deviations do not cancel out across positions. By applying the forward KL at every step, the objective also exhibits a mass-covering effect, discouraging mode collapse of the reference distribution while preserving diversity, in conjunction with the preference term $\mathcal{L}_{\text{pref}}$, which drives overall preference alignment.

\section{Experiments}
\label{sec:experiments}

\begin{table*}[t]
  \caption{Performance changes relative to the Base model in percentage points (pp, higher is better). Evaluated using LLM-as-Judge (LAJ-MQM) pass rates on 5,000-segment test sets for En$\to$Hi and En$\to$Tr, and human MQM pass rates on a 2,000-segment En$\to$Hi subset (24,446 words). LAJ-MQM values are mean $\pm$ standard deviation over five independent inference runs of a single training run.}
  \label{tab:main-results}
  \centering
  
  \resizebox{\textwidth}{!}{%
  \begin{tabular}{@{} c l *{8}{c} @{}}
    \toprule
    
    & \multirow{2}{*}{\textbf{Method}} & \multirow{2}{*}{\textbf{Verity}} & \multirow{2}{*}{\textbf{Design}} & \multirow{2}{*}{\textbf{Term.}} & \textbf{Locale} & \multirow{2}{*}{\textbf{Style}} & \multirow{2}{*}{\textbf{Acc.}} & \multirow{2}{*}{\textbf{Flu.}} & \multirow{2}{*}{\textbf{All-Pass}} \\
    & & & & & \textbf{Conv.} & & & & \\
    \midrule
    
    \multirow{7}{*}{\begin{tabular}{@{}c@{}}\textbf{En $\to$ Hi} \\ \textbf{(LAJ-MQM)}\end{tabular}} 
    & SFT & $-$1.7{\scriptsize$\pm$0.2} & $-$0.8{\scriptsize$\pm$0.2} & $-$3.3{\scriptsize$\pm$0.4} & $-$0.5{\scriptsize$\pm$0.3} & 2.0{\scriptsize$\pm$0.6} & 1.5{\scriptsize$\pm$0.3} & 12.1{\scriptsize$\pm$1.3} & 1.8{\scriptsize$\pm$0.6} \\
    & DPO & $-$0.1{\scriptsize$\pm$0.4} & $-$0.1{\scriptsize$\pm$0.3} & $-$0.7{\scriptsize$\pm$0.2} & $-$0.5{\scriptsize$\pm$0.1} & 1.0{\scriptsize$\pm$0.2} & $-$0.3{\scriptsize$\pm$0.5} & 0.7{\scriptsize$\pm$0.4} & $-$0.4{\scriptsize$\pm$1.3} \\
    & APO-Down & $-$0.1{\scriptsize$\pm$0.1} & $-$0.5{\scriptsize$\pm$0.4} & $-$1.0{\scriptsize$\pm$0.3} & $-$0.5{\scriptsize$\pm$0.2} & 2.3{\scriptsize$\pm$1.3} & $-$1.2{\scriptsize$\pm$0.7} & 0.2{\scriptsize$\pm$0.5} & 0.6{\scriptsize$\pm$0.4} \\
    & TDPO & 0.1{\scriptsize$\pm$0.3} & $-$0.2{\scriptsize$\pm$0.8} & $-$0.3{\scriptsize$\pm$0.2} & $-$0.4{\scriptsize$\pm$0.2} & 1.4{\scriptsize$\pm$0.9} & 0.5{\scriptsize$\pm$0.1} & 1.8{\scriptsize$\pm$2.4} & 0.5{\scriptsize$\pm$1.4} \\
    & BAPO & $-$0.1{\scriptsize$\pm$0.3} & $-$0.1{\scriptsize$\pm$0.3} & $-$0.6{\scriptsize$\pm$0.3} & $-$0.5{\scriptsize$\pm$0.2} & 1.2{\scriptsize$\pm$1.6} & $-$0.4{\scriptsize$\pm$0.5} & 0.9{\scriptsize$\pm$1.5} & $-$0.2{\scriptsize$\pm$0.8} \\
    & BAPO-Down & $-$0.2{\scriptsize$\pm$0.2} & $-$0.3{\scriptsize$\pm$0.3} & $-$1.2{\scriptsize$\pm$0.4} & $-$0.7{\scriptsize$\pm$0.2} & 15.4{\scriptsize$\pm$0.8} & $-$6.4{\scriptsize$\pm$1.2} & \textbf{16.9}{\scriptsize$\pm$1.8} & 7.4{\scriptsize$\pm$0.8} \\
    & \textbf{StalePO} & $-$0.5{\scriptsize$\pm$0.4} & $-$1.1{\scriptsize$\pm$0.3} & $-$2.0{\scriptsize$\pm$0.5} & $-$0.6{\scriptsize$\pm$0.2} & \textbf{20.9}{\scriptsize$\pm$0.7} & $-$5.4{\scriptsize$\pm$0.1} & 16.5{\scriptsize$\pm$0.9} & \textbf{14.9}{\scriptsize$\pm$0.4} \\
    \midrule
    
    \multirow{2}{*}{\begin{tabular}{@{}c@{}}\textbf{En $\to$ Tr} \\ \textbf{(LAJ-MQM)}\end{tabular}}
    & DPO & $-$0.1{\scriptsize$\pm$0.5} & $-$0.2{\scriptsize$\pm$0.3} & $-$0.3{\scriptsize$\pm$0.6} & $-$0.1{\scriptsize$\pm$0.1} & 0.7{\scriptsize$\pm$0.2} & $-$0.7{\scriptsize$\pm$0.3} & 0.4{\scriptsize$\pm$1.8} & $-$0.2{\scriptsize$\pm$1.3}\\
    & \textbf{StalePO} & 0.0{\scriptsize$\pm$0.2} & 0.1{\scriptsize$\pm$0.3} & 1.6{\scriptsize$\pm$0.3} & 0.3{\scriptsize$\pm$0.2} & \textbf{2.3}{\scriptsize$\pm$1.1} & 1.0{\scriptsize$\pm$0.1} & 1.3{\scriptsize$\pm$1.0} & \textbf{4.6}{\scriptsize$\pm$0.4} \\
    \midrule
    
    \begin{tabular}{@{}c@{}}\textbf{En $\to$ Hi} \\ \textbf{(Human)}\end{tabular}
    & \textbf{StalePO} & 0.0 & 0.0 & 0.8 & 0.0 & \textbf{15.9} & $-$3.5 & \textbf{37.3} & \textbf{13.8} \\
    
    \bottomrule
  \end{tabular}%
  }
\end{table*}

\subsection{Evaluation}

Our primary evaluation uses a seven-category automated Multidimensional Quality Metrics (MQM) framework \citep{lommel2014mqm, freitag2021experts} with LLM-as-Judge (LAJ-MQM)-based scorers optimized for each dimension: Verity (legality), Design (markup integrity), Terminology (termbase compliance), Locale Convention (formatting and units), Style (company style adherence), Accuracy (additions, omissions, mistranslations), and Fluency (punctuation, grammar). Each scorer produces a binary pass/fail judgment per segment. All scorers are implemented with Claude Sonnet 4.5, a model whose family differs from that of the GPT-OSS policy under evaluation. This cross-family setup mitigates the self-preference bias that can arise when a model judges outputs sampled from its own family. We define the \textbf{all-pass} rate as the fraction of segments passing all seven scorers simultaneously and adopt it as our primary aggregate metric.

Reference-based metrics are unsuitable in this regime. A correct conversational transliteration that diverges from the post-edit improves translation quality while lowering BLEU and raising TER, whereas LAJ-MQM scores translations independently of the post-edits. To validate the LAJ-MQM judgments, professional translators annotate a 2,000-segment English$\rightarrow$Hindi subset under the same seven-category taxonomy. The resulting pass-rate changes (Table~\ref{tab:main-results}) agree with LAJ-MQM in both direction and concentration, with the largest gains in Style and Fluency and a small Accuracy regression.

\subsection{Setup}

\paragraph{Data} We begin with a large corpus of post-edits on English$\rightarrow$Hindi and English$\rightarrow$Turkish machine translations produced by previous generations of NMT systems. For English$\rightarrow$Hindi data, we apply a Kimi K2.5-based classifier to retain post-edits exhibiting stylistic modernization (formal$\rightarrow$conversational lexical shifts, passive$\rightarrow$active simplification) and punctuation normalization. This yields 24k train / 633 validation segments, concentrating supervision on style and fluency. In the notation of Section~\ref{sec:stale_preference}, this filtering retains examples whose advantage subset $\mathcal{S}$ is concentrated on style and fluency, whereas the unfiltered English$\rightarrow$Turkish setting below leaves $\mathcal{S}$ unconstrained. In the absence of ground-truth style labels, we omit precision/recall and evaluate filtering via downstream gains (Section~\ref{sec:results}).

For English$\rightarrow$Turkish, we retain all post-edits without filtering, yielding an unfiltered dataset of 10k train / 598 validation segments, serving as a complementary setting.
 
Each training example consists of a source prompt $x$ (English text augmented with a target-language style guide and termbase entries), a preferred response $y_w$ (human post-edit of the NMT output), a dispreferred response $y_l$ (raw NMT output), and a base response $y_b$ (sampled from GPT-OSS 120B at temperature 1.0, top-$p$ 1.0). We evaluate on stratified, temporally disjoint test sets of 5,000 segments per language pair. All results are reported across the seven MQM categories to capture cross-dimensional trade-offs.

\textbf{Model and Training.} We fine-tune a 120B-parameter GPT-OSS model using QLoRA \citep{dettmers2024qlora} adapters on all linear projection layers. All methods are trained for 4 epochs with early stopping on a single node of 8 NVIDIA A100 GPUs using AdamW, with an effective batch size of 64, learning rate $3 \times 10^{-6}$, cosine decay, and LoRA rank $r=8$. Hyperparameter configurations and early-stopping checkpoints for every method are selected exclusively on the corresponding validation set. The test sets are used only for final reporting. Every baseline is tuned independently over its own sweep of $\beta$ and any coefficients specific to that method. At inference, we decode with temperature 1.0 and top-$p$ 1.0.

\begin{figure*}[t]
\centering
\includegraphics[width=0.9\linewidth]{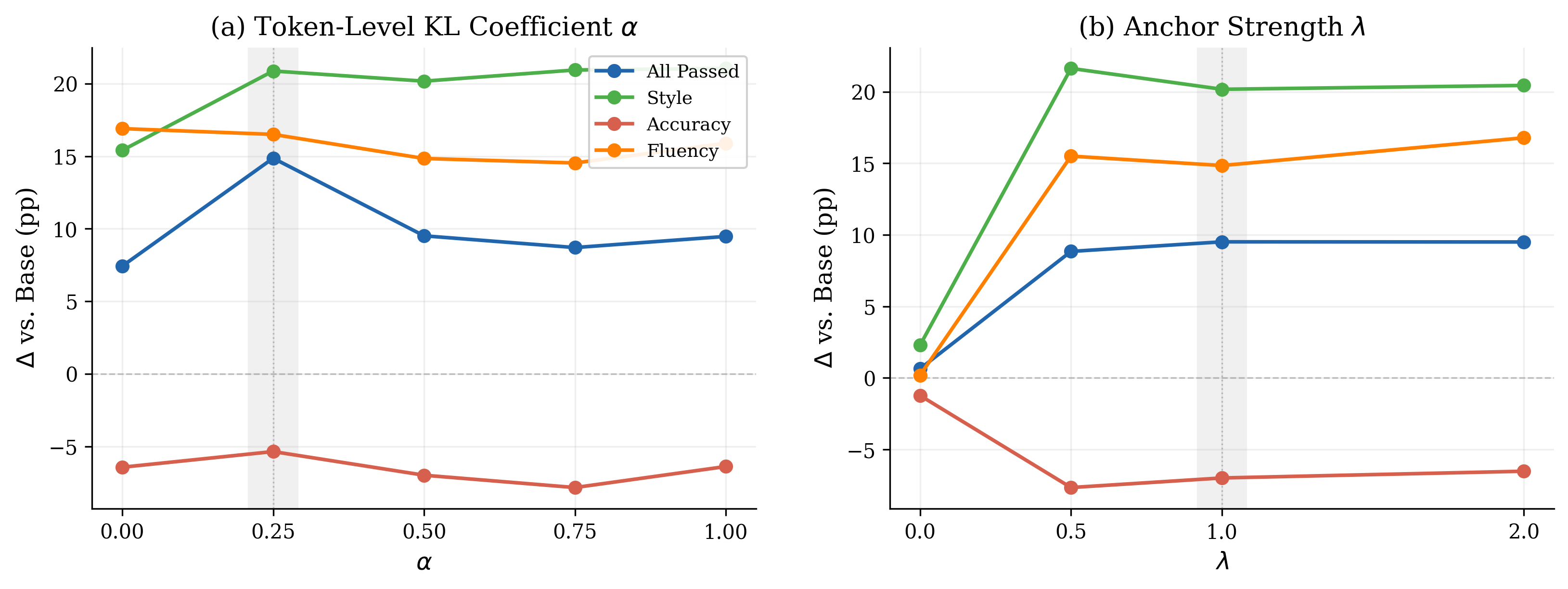}
\caption{
StalePO hyperparameter sensitivity: pass-rate change vs.\ base. (a)~Token-level KL coefficient $\alpha$ ($\lambda{=}1.0$, $r{=}8$).
(b)~Anchor strength $\lambda$ ($\alpha{=}0.5$, $r{=}8$).}
\label{fig:hparam-ablation}
\end{figure*}

\begin{table}[htbp]
  \caption{Component composition of each method. Down: downward directional control; Anchor: base-response anchor; TokenKL: token-level sequential KL; SFT($y_w$): supervised fine-tuning on post-edits.}
  \label{tab:ablation-matrix}
  \centering\small
  \begin{tabular}{@{}lcccc@{}}
    \toprule
    Method & SFT($y_w$) & Down & Anchor & TokenKL \\
    \midrule
    SFT              & \checkmark & -  & -  & -  \\
    DPO              & -  & -  & -  & -  \\
    TDPO             & -  & -  & -  & \checkmark \\
    BAPO             & -  & -  & \checkmark & -  \\
    APO-Down         & -  & \checkmark & -  & -  \\
    BAPO-Down        & -  & \checkmark & \checkmark & -  \\
    \textbf{StalePO} & -  & \checkmark & \checkmark & \checkmark \\
    \bottomrule
  \end{tabular}
\end{table}

\textbf{Baselines and Ablations.} Each baseline removes one or more of StalePO's three components. \textbf{SFT} fine-tunes the base model on the post-edits $y_w$ alone, representing the ``imitate-the-winner'' strategy and the degenerate case of likelihood-maximizing MT objectives such as CPO \citep{xu2024cpo}. \textbf{DPO} \citep{rafailov2023direct} applies the standard sigmoid objective. \textbf{APO-Down} \citep{doosterlinck2025anchored} adds the downward term only. \textbf{TDPO} \citep{zeng2024token} adds the token-level sequential KL. \textbf{BAPO} \citep{lee2024bapo} adds the base anchor. \textbf{BAPO-Down} uses the downward term with the base anchor, that is, StalePO without the token-level KL. Table~\ref{tab:ablation-matrix} summarizes the composition of each method.

\section{Results}
\label{sec:results}

On English$\rightarrow$Hindi (Table~\ref{tab:main-results}), StalePO achieves an improvement on all-pass rate of \textbf{14.9\,pp} over the base model under LAJ-MQM. Human MQM shows a \textbf{13.8\,pp} increase in all-pass rate, confirming consistent quality gains. These improvements are mainly driven by Style and Fluency, which increase by 20.9\,pp and 16.5\,pp, with corresponding human MQM pass-rate gains of 15.9\,pp and 37.3\,pp. Other dimensions remain largely stable (within $\pm$1--2\,pp). Qualitatively, the shift is a consistent move from formal Sanskrit-derived vocabulary toward the conversational transliterations mandated by the style guide. The only notable exception is Accuracy under LAJ-MQM, which decreases by 5.4\,pp. This is consistent with the stale preference formulation. Gains concentrate on the small subset $\mathcal{S}$ of dimensions where the post-edits genuinely improve on the current model, here style and fluency. Concretely, the model oversimplifies translations where simplification is unwarranted, leading to degradation in the Accuracy dimension. A human MQM subcategory breakdown localizes most of this regression to minor-severity untranslated spans, while mistranslation and omission counts remain nearly unchanged.

On English$\rightarrow$Turkish, the base model starts from a substantially stronger baseline, reflecting fewer style inconsistencies. Despite this stronger starting point, StalePO further improves performance by \textbf{4.6\,pp}. Similar to English$\rightarrow$Hindi, gains are concentrated in Style (+2.3\,pp) and Fluency (+1.3\,pp), while other dimensions remain stable or improve marginally. The gains are smaller and more diffuse than in the filtered English$\rightarrow$Hindi setting, as expected when $\mathcal{S}$ is not concentrated on a single dimension by construction. This consistency across language pairs suggests that StalePO generalizes across typologically distinct settings and varying baseline quality regimes, while naive DPO fails in both.

\begin{figure*}[t]
\centering
\includegraphics[width=\linewidth]{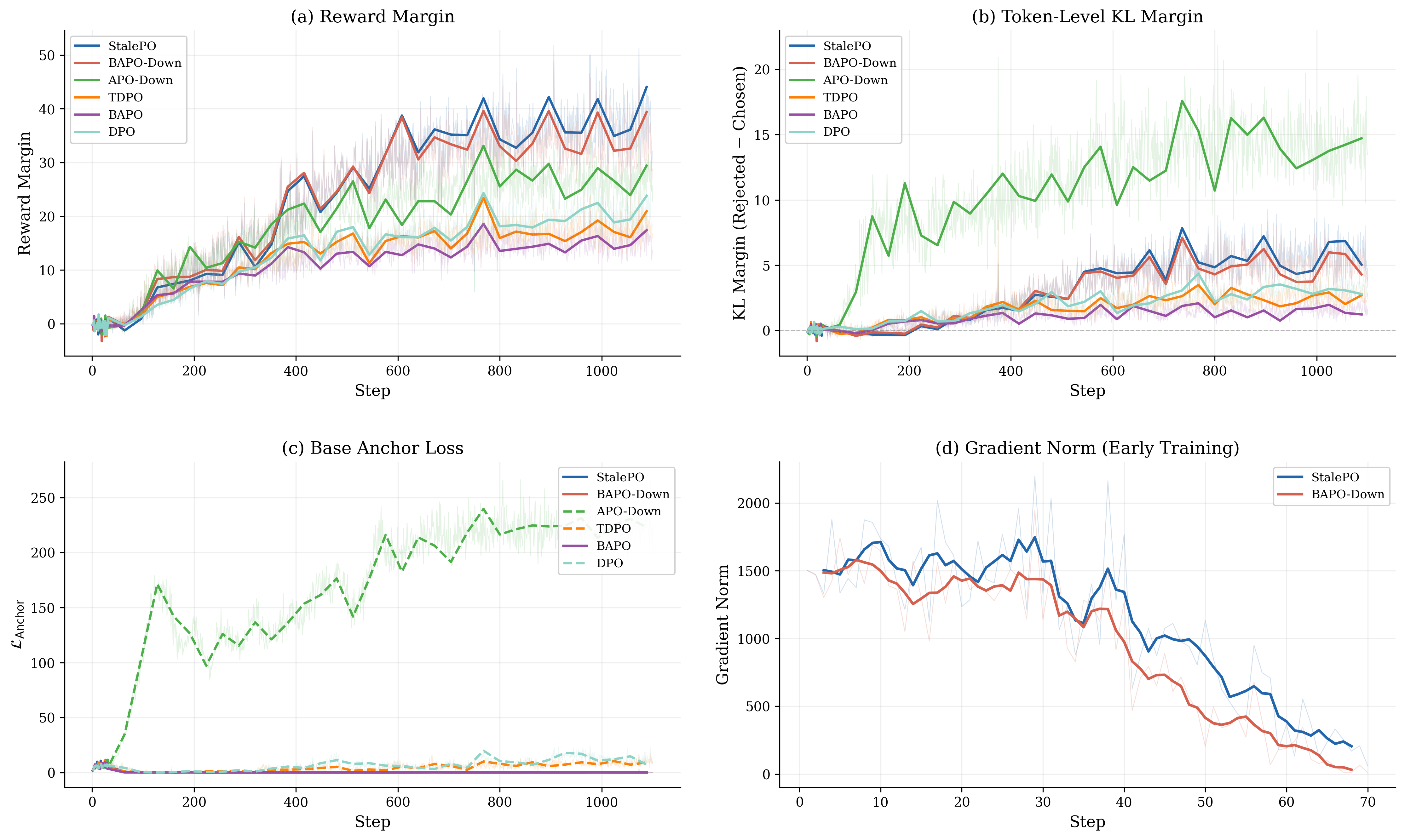}
\caption{
Training dynamics across ablation methods.
(a) Reward margin.
(b) Token-level KL margin (rejected minus chosen).
(c) Base anchor loss $\mathcal{L}_{\text{Anchor}}$.
(d) Gradient norm during early training (first 70 steps).
}
\label{fig:training-dynamics}
\end{figure*}

\subsection{Baseline Ablation}
\label{sec:baseline-ablation}

Neither imitating the post-edits nor any single component of StalePO improves on the base model. Supervised fine-tuning on the post-edits alone (SFT) is a particularly informative baseline: despite the En$\to$Hi training data being explicitly filtered to concentrate on stylistic modernization, SFT improves Style by only +2.0\,pp and the all-pass rate by only +1.8\,pp, even as it raises Fluency (+12.1\,pp). In other words, directly imitating the post-edits fails to internalize the very style signal the data was curated to emphasize, confirming that the stale post-edits are not a target the model should imitate. 

Methods carrying a single component perform comparably to the base model. Tuned DPO shifts the all-pass rate by only $-$0.4\,pp, TDPO by 0.5\,pp, and BAPO by $-$0.2\,pp, indicating that anchoring or an alternative optimization objective alone does not convert a stale preference signal into improved translation quality. Directional control in isolation is similarly insufficient. APO-Down improves by only 0.6\,pp, and its run diverges further from the base response distribution than any anchored variant, consistent with unconstrained downward updates eroding the model's original response distribution. A plausible explanation is that, without explicit anchoring, optimization amplifies biases in sampled outputs and drifts away from the distribution the model started from.

Introducing a base-response anchor leads to substantial improvements, with BAPO-Down improving all-pass by 7.4\,pp, alongside gains in Style (15.4\,pp) and Fluency (16.9\,pp), though at the cost of reduced Accuracy ($-$6.4\,pp). Building on this, StalePO attains the best overall performance of +14.9\,pp, further improving Style (20.9\,pp) while partially recovering Accuracy ($-$5.4\,pp). These gains stem from the per-position divergence constraint, which limits updates on already-diverged tokens and encourages more localized, stable adaptations. Overall, the results show that anchoring, directional guidance, and token-level constraints are jointly necessary to incorporate preference signals while maintaining response fidelity.

\subsection{Hyperparameter Ablation}
\label{sec:hparam-ablation}

We perform an ablation study on two hyperparameters: the anchor strength $\lambda \in \{0.5, 1.0, 2.0\}$, which balances preservation of the model's base behavior against learning from the preference signal, and the token-level KL coefficient $\alpha \in \{0.25, 0.5, 0.75, 1.0\}$, which modulates per-position divergence attenuation in the sigmoid saturation regime. We vary each parameter independently while holding the other fixed: the $\alpha$ sweep uses $\lambda = 1.0$, and the $\lambda$ sweep uses $\alpha = 0.5$. Our main configuration (Table~\ref{tab:main-results}) uses $\lambda = 1.0$, $\alpha = 0.25$, which corresponds to the peak of the $\alpha$ sweep. Figure~\ref{fig:hparam-ablation} reports the resulting change in pass rate relative to the base model.

\paragraph{Effect of $\alpha$.} As shown in Figure~\ref{fig:hparam-ablation}(a), the all-pass rate peaks at $\alpha = 0.25$ and decreases for higher values, though all configurations remain substantially above the base model. 

\paragraph{Effect of $\lambda$.} The anchor's effect is pronounced (Figure~\ref{fig:hparam-ablation}(b)): without it, all metrics remain near the base model despite the intended directional bias. Introducing even a modest anchor ($\lambda = 0.5$) produces a substantial improvement across all dimensions, although the effect plateaus at higher strengths. This behavior is consistent with the observation of \citet{lee2024bapo} that anchoring effectiveness saturates beyond a threshold.

\subsection{Training Dynamics}
\label{sec:training-dynamics}

The optimization behavior underlying these outcomes is visible in the training dynamics. Figure~\ref{fig:training-dynamics} tracks reward margins, token-level divergences, the base anchor loss, and gradient norms across methods.

\paragraph{Directional Signal and Anchor Stability} APO-Down produces larger reward margins than the DPO variants under stale preferences (Figure~\ref{fig:training-dynamics}(a)), indicating a stronger directional signal, but at the cost of stability. Without an anchor it diverges progressively from the base response distribution throughout training (Figure~\ref{fig:training-dynamics}(c)), degrading generation quality. The anchor resolves this: StalePO and BAPO-Down both satisfy the constraint within the first seventy steps and train stably thereafter, ending two orders of magnitude below the unanchored runs on $\mathcal{L}_{\text{Anchor}}$. Sigmoid DPO methods drift moderately without the anchor but do not collapse, consistent with their inherently more conservative gradient dynamics.

\paragraph{Asymmetric Divergence} Among the anchored directional variants, StalePO maintains a slightly larger late-training KL margin than BAPO-Down (Figure~\ref{fig:training-dynamics}(b)), indicating stronger relative divergence along rejected trajectories. Their gradient norms begin at a comparable scale, but BAPO-Down's declines faster while StalePO retains a learning signal through early training (Figure~\ref{fig:training-dynamics}(d)). Together, these patterns are consistent with the token-level KL sustaining asymmetric adaptation after the anchor has stabilized.

\section{Conclusion}

This work formalizes the stale preference problem in machine translation and proposes StalePO, an objective derived from the constraints of this setting. StalePO improves the proportion of error-free segments by 14.9\,pp on English$\rightarrow$Hindi and 4.6\,pp on English$\rightarrow$Turkish, with the largest gains in style adherence and fluency, and a human MQM evaluation corroborates the English$\rightarrow$Hindi result at 13.8\,pp.

Moreover, the stale preference problem is not specific to machine translation. It arises whenever model updates outpace the collection of fresh preference data. StalePO mitigates this issue by learning to avoid failure modes in stale preferences rather than imitating them. Extending this approach to other domains and to richer, multi-error preference data is a promising direction for future work.

\section*{Limitations}

Our work has a few limitations. First, due to budget constraints, professional human MQM annotation is not feasible at the scale of our test sets, so our primary evaluation relies on an LLM-as-Judge MQM framework. We validate it against human annotators on an English$\rightarrow$Hindi subset and use a cross-family judge to limit self-preference bias, but our English$\rightarrow$Turkish results are not backed by human evaluation. Relatedly, company policy prevents us from disclosing absolute MQM pass rates, so all pass-rate results are reported as changes relative to the base model. Second, StalePO offers no benefit when the updated model's outputs are already better than the post-edits on every dimension. In this case the advantage subset $\mathcal{S}$ is empty and the preference pair carries no signal to learn from, so the objective can only preserve existing behavior rather than improve it. Third, and conversely, StalePO is not designed for the standard setting where the post-edits are uniformly better than the model, because it biases both responses' likelihoods downward rather than imitating the preferred output. When the preferred output is a genuine quality target to approach, methods that imitate the winner are more appropriate.

\section*{Acknowledgments}

We sincerely thank our teammate Aiden Watson for his valuable support in facilitating access to the GPUs and other computational resources. His assistance was instrumental in enabling the experiments presented in this paper.

\bibliography{custom}

\begin{thebibliography}{32}
\providecommand{\natexlab}[1]{#1}

\bibitem[{Bai et~al.(2022)Bai, Jones, Ndousse, Askell, Chen, DasSarma, Drain,
  Fort, Ganguli, Henighan et~al.}]{bai2022training}
Yuntao Bai, Andy Jones, Kamal Ndousse, Amanda Askell, Anna Chen, Nova DasSarma,
  Dawn Drain, Stanislav Fort, Deep Ganguli, Tom Henighan, et~al. 2022.
\newblock Training a helpful and harmless assistant with reinforcement learning
  from human feedback.
\newblock \emph{arXiv preprint arXiv:2204.05862}.

\bibitem[{Berger et~al.(2024)Berger, Exel, Huck, and
  Riezler}]{berger2024postedits}
Nathaniel Berger, Miriam Exel, Matthias Huck, and Stefan Riezler. 2024.
\newblock \href {https://doi.org/10.18653/v1/2024.wmt-1.122} {Post-edits are
  preferences too}.
\newblock In \emph{Proceedings of the Ninth Conference on Machine Translation},
  pages 1289--1300, Miami, Florida, USA. Association for Computational
  Linguistics.

\bibitem[{Calandriello et~al.(2024)Calandriello, Guo, Munos, Rowland, Tang,
  Avila~Pires, Richemond, Le~Lan, Valko, Liu, Joshi, Zheng, and
  Piot}]{calandriello2024online}
Daniele Calandriello, Zhaohan~Daniel Guo, Remi Munos, Mark Rowland, Yunhao
  Tang, Bernardo Avila~Pires, Pierre~Harvey Richemond, Charline Le~Lan, Michal
  Valko, Tianqi Liu, Rishabh Joshi, Zeyu Zheng, and Bilal Piot. 2024.
\newblock \href {https://proceedings.mlr.press/v235/calandriello24a.html}
  {Human alignment of large language models through online preference
  optimisation}.
\newblock In \emph{Proceedings of the 41st International Conference on Machine
  Learning}, volume 235 of \emph{Proceedings of Machine Learning Research},
  pages 5409--5435. PMLR.

\bibitem[{Chen et~al.(2024{\natexlab{a}})Chen, He, Yuan, Cui, Su, and
  Zhu}]{chen2024nca}
Huayu Chen, Guande He, Lifan Yuan, Ganqu Cui, Hang Su, and Jun Zhu.
  2024{\natexlab{a}}.
\newblock \href {https://doi.org/10.52202/079017-3741} {Noise contrastive
  alignment of language models with explicit rewards}.
\newblock In \emph{Advances in Neural Information Processing Systems},
  volume~37, pages 117784--117812. Curran Associates, Inc.

\bibitem[{Chen et~al.(2024{\natexlab{b}})Chen, Deng, Yuan, Ji, and
  Gu}]{chen2024spin}
Zixiang Chen, Yihe Deng, Huizhuo Yuan, Kaixuan Ji, and Quanquan Gu.
  2024{\natexlab{b}}.
\newblock \href {https://proceedings.mlr.press/v235/chen24j.html} {Self-play
  fine-tuning converts weak language models to strong language models}.
\newblock In \emph{Proceedings of the 41st International Conference on Machine
  Learning}, volume 235 of \emph{Proceedings of Machine Learning Research},
  pages 6621--6642. PMLR.

\bibitem[{Christiano et~al.(2017)Christiano, Leike, Brown, Martic, Legg, and
  Amodei}]{christiano2017deep}
Paul~F. Christiano, Jan Leike, Tom~B. Brown, Miljan Martic, Shane Legg, and
  Dario Amodei. 2017.
\newblock Deep reinforcement learning from human preferences.
\newblock In \emph{Proceedings of the 31st International Conference on Neural
  Information Processing Systems}, NIPS'17, page 4302–4310, Red Hook, NY,
  USA. Curran Associates Inc.

\bibitem[{Christopoulou et~al.(2025)Christopoulou, Cardenas, Lampouras,
  Bou~Ammar, and Wang}]{deng2024sparsepo}
Fenia Christopoulou, Ronald Cardenas, Gerasimos Lampouras, Haitham Bou~Ammar,
  and Jun Wang. 2025.
\newblock \href {https://doi.org/10.18653/v1/2025.findings-emnlp.1389}
  {{S}parse{PO}: Controlling preference alignment of {LLM}s via sparse token
  masks}.
\newblock In \emph{Findings of the Association for Computational Linguistics:
  EMNLP 2025}, pages 25477--25503, Suzhou, China. Association for Computational
  Linguistics.

\bibitem[{Dettmers et~al.(2023)Dettmers, Pagnoni, Holtzman, and
  Zettlemoyer}]{dettmers2024qlora}
Tim Dettmers, Artidoro Pagnoni, Ari Holtzman, and Luke Zettlemoyer. 2023.
\newblock \href
  {https://proceedings.neurips.cc/paper_files/paper/2023/file/1feb87871436031bdc0f2beaa62a049b-Paper-Conference.pdf}
  {Qlora: Efficient finetuning of quantized llms}.
\newblock In \emph{Advances in Neural Information Processing Systems},
  volume~36, pages 10088--10115. Curran Associates, Inc.

\bibitem[{D'Oosterlinck et~al.(2025)D'Oosterlinck, Xu, Develder, Demeester,
  Singh, Potts, Kiela, and Mehri}]{doosterlinck2025anchored}
Karel D'Oosterlinck, Winnie Xu, Chris Develder, Thomas Demeester, Amanpreet
  Singh, Christopher Potts, Douwe Kiela, and Shikib Mehri. 2025.
\newblock \href {https://doi.org/10.1162/tacl_a_00748} {Anchored preference
  optimization and contrastive revisions: Addressing underspecification in
  alignment}.
\newblock \emph{Transactions of the Association for Computational Linguistics},
  13:442--460.

\bibitem[{Ethayarajh et~al.(2024)Ethayarajh, Xu, Muennighoff, Jurafsky, and
  Kiela}]{ethayarajh2024kto}
Kawin Ethayarajh, Winnie Xu, Niklas Muennighoff, Dan Jurafsky, and Douwe Kiela.
  2024.
\newblock \href {https://proceedings.mlr.press/v235/ethayarajh24a.html} {Model
  alignment as prospect theoretic optimization}.
\newblock In \emph{Proceedings of the 41st International Conference on Machine
  Learning}, volume 235 of \emph{Proceedings of Machine Learning Research},
  pages 12634--12651. PMLR.

\bibitem[{Freitag et~al.(2021)Freitag, Foster, Grangier, Ratnakar, Tan, and
  Macherey}]{freitag2021experts}
Markus Freitag, George Foster, David Grangier, Viresh Ratnakar, Qijun Tan, and
  Wolfgang Macherey. 2021.
\newblock \href {https://doi.org/10.1162/tacl_a_00437} {Experts, errors, and
  context: A large-scale study of human evaluation for machine translation}.
\newblock \emph{Transactions of the Association for Computational Linguistics},
  9:1460--1474.

\bibitem[{Gheshlaghi~Azar et~al.(2024)Gheshlaghi~Azar, Daniel~Guo, Piot, Munos,
  Rowland, Valko, and Calandriello}]{azar2024general}
Mohammad Gheshlaghi~Azar, Zhaohan Daniel~Guo, Bilal Piot, Remi Munos, Mark
  Rowland, Michal Valko, and Daniele Calandriello. 2024.
\newblock \href {https://proceedings.mlr.press/v238/gheshlaghi-azar24a.html} {A
  general theoretical paradigm to understand learning from human preferences}.
\newblock In \emph{Proceedings of The 27th International Conference on
  Artificial Intelligence and Statistics}, volume 238 of \emph{Proceedings of
  Machine Learning Research}, pages 4447--4455. PMLR.

\bibitem[{Guo et~al.(2024)Guo, Zhang, Liu, Liu, Khalman, Llinares, Ber
  et~al.}]{guo2024oaif}
Shangmin Guo, Biao Zhang, Tianlin Liu, Tianqi Liu, Misha Khalman, Felipe
  Llinares, Alexandre Ber, et~al. 2024.
\newblock Direct language model alignment from online {AI} feedback.
\newblock \emph{arXiv preprint arXiv:2402.04792}.

\bibitem[{Hong et~al.(2024)Hong, Lee, and Thorne}]{hong2024orpo}
Jiwoo Hong, Noah Lee, and James Thorne. 2024.
\newblock \href {https://doi.org/10.18653/v1/2024.emnlp-main.626} {{ORPO}:
  Monolithic preference optimization without reference model}.
\newblock In \emph{Proceedings of the 2024 Conference on Empirical Methods in
  Natural Language Processing}, pages 11170--11189, Miami, Florida, USA.
  Association for Computational Linguistics.

\bibitem[{Lee et~al.(2024)Lee, Jeong, Kim, Jung, Oh, Kim, and
  Yun}]{lee2024bapo}
Gihun Lee, Minchan Jeong, Yujin Kim, Hojung Jung, Jaehoon Oh, SangMook Kim, and
  Se-Young Yun. 2024.
\newblock \href {https://doi.org/10.18653/v1/2024.findings-emnlp.398} {{BAPO}:
  Base-anchored preference optimization for overcoming forgetting in large
  language models personalization}.
\newblock In \emph{Findings of the Association for Computational Linguistics:
  EMNLP 2024}, pages 6804--6820, Miami, Florida, USA. Association for
  Computational Linguistics.

\bibitem[{Liu et~al.(2024)Liu, Zhao, Joshi, Khalman, Saleh, Liu, and
  Liu}]{liu2024rso}
Tianqi Liu, Yao Zhao, Rishabh Joshi, Misha Khalman, Mohammad Saleh, Peter~J
  Liu, and Jialu Liu. 2024.
\newblock \href {https://openreview.net/forum?id=xbjSwwrQOe} {Statistical
  rejection sampling improves preference optimization}.
\newblock In \emph{The Twelfth International Conference on Learning
  Representations}.

\bibitem[{Lommel et~al.(2014)Lommel, Uszkoreit, and Burchardt}]{lommel2014mqm}
Arle Lommel, Hans Uszkoreit, and Aljoscha Burchardt. 2014.
\newblock \href {https://doi.org/10.5565/rev/tradumatica.77} {Multidimensional
  quality metrics (mqm): A framework for declaring and describing translation
  quality metrics}.
\newblock \emph{Tradumàtica tecnologies de la traducció}, pages 455--463.

\bibitem[{Meng et~al.(2024)Meng, Xia, and Chen}]{meng2024simpo}
Yu~Meng, Mengzhou Xia, and Danqi Chen. 2024.
\newblock Simpo: simple preference optimization with a reference-free reward.
\newblock In \emph{Proceedings of the 38th International Conference on Neural
  Information Processing Systems}, NIPS '24, Red Hook, NY, USA. Curran
  Associates Inc.

\bibitem[{Ouyang et~al.(2022)Ouyang, Wu, Jiang, Almeida, Wainwright, Mishkin,
  Zhang, Agarwal, Slama, Ray, Schulman, Hilton, Kelton, Miller, Simens, Askell,
  Welinder, Christiano, Leike, and Lowe}]{ouyang2022training}
Long Ouyang, Jeff Wu, Xu~Jiang, Diogo Almeida, Carroll~L. Wainwright, Pamela
  Mishkin, Chong Zhang, Sandhini Agarwal, Katarina Slama, Alex Ray, John
  Schulman, Jacob Hilton, Fraser Kelton, Luke Miller, Maddie Simens, Amanda
  Askell, Peter Welinder, Paul Christiano, Jan Leike, and Ryan Lowe. 2022.
\newblock Training language models to follow instructions with human feedback.
\newblock In \emph{Proceedings of the 36th International Conference on Neural
  Information Processing Systems}, NIPS '22, Red Hook, NY, USA. Curran
  Associates Inc.

\bibitem[{Pal et~al.(2024)Pal, Karkhanis, Dooley, Roberts, Naidu, and
  White}]{pal2024smaug}
Arka Pal, Deep Karkhanis, Samuel Dooley, Manley Roberts, Siddartha Naidu, and
  Colin White. 2024.
\newblock Smaug: Fixing failure modes of preference optimisation with
  dpo-positive.
\newblock \emph{arXiv preprint arXiv:2402.13228}.

\bibitem[{Park et~al.(2024)Park, Rafailov, Ermon, and
  Finn}]{park2024disentangling}
Ryan Park, Rafael Rafailov, Stefano Ermon, and Chelsea Finn. 2024.
\newblock \href {https://doi.org/10.18653/v1/2024.findings-acl.297}
  {Disentangling length from quality in direct preference optimization}.
\newblock In \emph{Findings of the Association for Computational Linguistics:
  ACL 2024}, pages 4998--5017, Bangkok, Thailand. Association for Computational
  Linguistics.

\bibitem[{Rafailov et~al.(2023)Rafailov, Sharma, Mitchell, Ermon, Manning, and
  Finn}]{rafailov2023direct}
Rafael Rafailov, Archit Sharma, Eric Mitchell, Stefano Ermon, Christopher~D.
  Manning, and Chelsea Finn. 2023.
\newblock Direct preference optimization: your language model is secretly a
  reward model.
\newblock In \emph{Proceedings of the 37th International Conference on Neural
  Information Processing Systems}, NIPS '23, Red Hook, NY, USA. Curran
  Associates Inc.

\bibitem[{Razin et~al.(2024)Razin, Malladi, Bhaskar, Chen, Arora, and
  Hanin}]{tajwar2025preference}
Noam Razin, Sadhika Malladi, Adithya Bhaskar, Danqi Chen, Sanjeev Arora, and
  Boris Hanin. 2024.
\newblock Unintentional unalignment: Likelihood displacement in direct
  preference optimization.
\newblock \emph{arXiv preprint arXiv:2410.08847}.

\bibitem[{Schulman et~al.(2017)Schulman, Wolski, Dhariwal, Radford, and
  Klimov}]{schulman2017proximal}
John Schulman, Filip Wolski, Prafulla Dhariwal, Alec Radford, and Oleg Klimov.
  2017.
\newblock Proximal policy optimization algorithms.
\newblock \emph{arXiv preprint arXiv:1707.06347}.

\bibitem[{Wu et~al.(2025)Wu, Sun, Yuan, Ji, Yang, and Gu}]{wu2025sppo}
Yue Wu, Zhiqing Sun, Huizhuo Yuan, Kaixuan Ji, Yiming Yang, and Quanquan Gu.
  2025.
\newblock \href {https://openreview.net/forum?id=a3PmRgAB5T} {Self-play
  preference optimization for language model alignment}.
\newblock In \emph{The Thirteenth International Conference on Learning
  Representations}.

\bibitem[{Xu et~al.(2024)Xu, Sharaf, Chen, Tan, Shen, Van~Durme, Murray, and
  Kim}]{xu2024cpo}
Haoran Xu, Amr Sharaf, Yunmo Chen, Weiting Tan, Lingfeng Shen, Benjamin
  Van~Durme, Kenton Murray, and Young~Jin Kim. 2024.
\newblock \href {https://proceedings.mlr.press/v235/xu24t.html} {Contrastive
  preference optimization: Pushing the boundaries of {LLM} performance in
  machine translation}.
\newblock In \emph{Proceedings of the 41st International Conference on Machine
  Learning}, volume 235 of \emph{Proceedings of Machine Learning Research},
  pages 55204--55224. PMLR.

\bibitem[{Xu et~al.(2026)Xu, Vemuri, Panaganti, Kalathil, Jain, and
  Ramachandran}]{cheng2026drdpo}
Zaiyan Xu, Sushil Vemuri, Kishan Panaganti, Dileep Kalathil, Rahul Jain, and
  Deepak Ramachandran. 2026.
\newblock \href {https://openreview.net/forum?id=D19hc2XPeZ} {Robust {LLM}
  alignment via distributionally robust direct preference optimization}.
\newblock In \emph{The Thirty-ninth Annual Conference on Neural Information
  Processing Systems}.

\bibitem[{Yang et~al.(2025)Yang, Liu, Xie, Huang, Min, and
  Ananiadou}]{cao2024sepo}
Kailai Yang, Zhiwei Liu, Qianqian Xie, Jimin Huang, Erxue Min, and Sophia
  Ananiadou. 2025.
\newblock Selective preference optimization via token-level reward function
  estimation.
\newblock In \emph{Proceedings of the 2025 Conference on Empirical Methods in
  Natural Language Processing}, pages 7043--7067.

\bibitem[{Yin et~al.(2025)Yin, Leong, Zhang, Zhu, Yan, Zhang, He, Li, Wang,
  Zhang, and Yang}]{zhang2024fpo}
Qingyu Yin, Chak~Tou Leong, Hongbo Zhang, Minjun Zhu, Hanqi Yan, Qiang Zhang,
  Yulan He, Wenjie Li, Jun Wang, Yue Zhang, and Linyi Yang. 2025.
\newblock \href {https://openreview.net/forum?id=BCKSxOFX85} {Constrain
  alignment with sparse autoencoders}.
\newblock In \emph{Forty-second International Conference on Machine Learning}.

\bibitem[{Zeng et~al.(2024)Zeng, Liu, Ma, Yang, Zhang, and
  Wang}]{zeng2024token}
Yongcheng Zeng, Guoqing Liu, Weiyu Ma, Ning Yang, Haifeng Zhang, and Jun Wang.
  2024.
\newblock Token-level direct preference optimization.
\newblock In \emph{Proceedings of the 41st International Conference on Machine
  Learning}, ICML'24. JMLR.org.

\bibitem[{Zhao et~al.(2023)Zhao, Khalman, Joshi, Narayan, Saleh, and
  Liu}]{zhao2023slic}
Yao Zhao, Mikhail Khalman, Rishabh Joshi, Shashi Narayan, Mohammad Saleh, and
  Peter~J Liu. 2023.
\newblock \href {https://openreview.net/forum?id=0qSOodKmJaN} {Calibrating
  sequence likelihood improves conditional language generation}.
\newblock In \emph{The Eleventh International Conference on Learning
  Representations}.

\bibitem[{Zhou et~al.(2025)Zhou, Zhang, Zhao, and Meng}]{zhou2024treg}
Wenxuan Zhou, Shujian Zhang, Lingxiao Zhao, and Tao Meng. 2025.
\newblock \href {https://doi.org/10.18653/v1/2025.acl-long.1353} {{T}-{REG}:
  Preference optimization with token-level reward regularization}.
\newblock In \emph{Proceedings of the 63rd Annual Meeting of the Association
  for Computational Linguistics (Volume 1: Long Papers)}, pages 27876--27889,
  Vienna, Austria. Association for Computational Linguistics.

\end{thebibliography}

\appendix

\section{Lexical Bias of Post-Edits Toward the Legacy NMT System}
\label{app:ter_bias}

We quantify the structural bias across the full English$\rightarrow$Hindi training set (24{,}220 segments) using Translation Edit Rate (TER). For each segment, we compute $\text{TER}(\text{NMT}, \text{PE})$ between the legacy NMT output and the human post-edit, and $\text{TER}(\text{Base}, \text{PE})$ between the GPT-OSS 120B output and the same post-edit. If post-editors produced translations independent of the NMT output, both scores should be comparable.

\begin{table}[htbp]
\centering
\resizebox{\linewidth}{!}{%
\begin{tabular}{@{}lcc@{}}
\toprule
& \textbf{NMT $\leftrightarrow$ Post-Edit} & \textbf{Base $\leftrightarrow$ Post-Edit} \\
\midrule
Mean TER   & 0.426 & 0.517 \\
Median TER & 0.379 & 0.500 \\
Segment-level proximity & 55.3\% & 23.7\% \\
\bottomrule
\end{tabular}%
}
\vspace{4pt}
\caption{TER between post-edits and each system on the En$\to$Hi training set. The remaining 21.0\% of segments are equidistant.}
\label{tab:ter-bias}
\end{table}

As shown in Table~\ref{tab:ter-bias}, post-edits are substantially closer to the NMT outputs (mean TER 0.426) than to the base model outputs (0.517), a difference of 0.091 (paired $t = 46.9$, $p \ll 0.001$). Segment-level proximity reports the fraction of segments for which $\text{TER}(\text{NMT}, \text{PE}) < \text{TER}(\text{Base}, \text{PE})$ and vice versa. Post-edits are 2.3$\times$ more likely to be closer to the NMT output than to the base model (55.3\% vs.\ 23.7\%). This confirms that post-editors correct rather than retranslate: their edits preserve the lexical scaffolding of the original NMT output.

\paragraph{Reference-based metrics invert quality rankings.} This bias propagates to evaluation when post-edits serve as references. Table~\ref{tab:ref-metrics} reports TER, COMET, chrF, and BLEU on the 5{,}000-segment test set, averaged over 5 runs.

SFT dominates the reference-based metrics, attaining the lowest TER and the highest chrF and BLEU, with BLEU rising 8.4 points above the base model. This is expected, since SFT is trained to reproduce the post-edits that serve as references here, which makes it close to an upper bound on agreement with them. Its all-pass rate nonetheless improves by only 1.8\,pp. DPO and TDPO also improve on all four metrics, with TDPO attaining the highest COMET, while leaving the all-pass rate essentially unchanged from the base model under LAJ-MQM. StalePO and BAPO-Down, the two methods with the largest all-pass gains (+14.9\,pp and +7.4\,pp), are the two worst on TER, and StalePO is worst of all on COMET, chrF, and BLEU. This inversion is expected: methods that produce higher-quality translations diverge from the NMT-anchored post-edits, incurring higher edit distances precisely because they generate superior translations. StalePO's conversational transliterations (e.g., \textit{\texthindi{एंटर}} for ``enter'') are penalized relative to the formal post-edit vocabulary (\textit{\texthindi{दर्ज}}), even though the transliterated forms are mandated by the style guide.

\begin{table}[htbp]
\centering
\caption{Reference-based metrics on the En$\to$Hi test set (5{,}000 segments), computed against post-edit references. Values are mean $\pm$ standard deviation over 5 runs.}
\label{tab:ref-metrics}

\resizebox{\columnwidth}{!}{%
\begin{tabular}{@{}lcccc@{}}
\toprule
\textbf{Method} & \textbf{TER} $\downarrow$ & \textbf{COMET} $\uparrow$ & \textbf{chrF} $\uparrow$ & \textbf{BLEU} $\uparrow$ \\
\midrule
Base      & 51.57{\scriptsize\,$\pm$0.48} & 83.59{\scriptsize\,$\pm$0.16} & 55.60{\scriptsize\,$\pm$0.31} & 37.41{\scriptsize\,$\pm$0.47} \\
SFT       & \textbf{50.18}{\scriptsize\,$\pm$0.13} & 83.65{\scriptsize\,$\pm$0.08} & \textbf{59.66}{\scriptsize\,$\pm$0.32} & \textbf{45.82}{\scriptsize\,$\pm$0.37} \\
DPO       & 50.35{\scriptsize\,$\pm$0.17} & 83.83{\scriptsize\,$\pm$0.23} & 56.35{\scriptsize\,$\pm$0.16} & 39.07{\scriptsize\,$\pm$0.37} \\
TDPO      & 50.51{\scriptsize\,$\pm$0.27} & 83.90{\scriptsize\,$\pm$0.19} & 56.15{\scriptsize\,$\pm$0.39} & 38.59{\scriptsize\,$\pm$0.33} \\
APO-Down  & 50.51{\scriptsize\,$\pm$0.51} & 83.70{\scriptsize\,$\pm$0.17} & 56.01{\scriptsize\,$\pm$0.52} & 38.94{\scriptsize\,$\pm$0.65} \\
BAPO      & 50.41{\scriptsize\,$\pm$0.25} & 83.84{\scriptsize\,$\pm$0.15} & 56.08{\scriptsize\,$\pm$0.23} & 39.05{\scriptsize\,$\pm$0.34} \\
BAPO-Down & 56.79{\scriptsize\,$\pm$0.11} & 83.10{\scriptsize\,$\pm$0.14} & 52.34{\scriptsize\,$\pm$0.26} & 41.04{\scriptsize\,$\pm$0.21} \\
StalePO   & 53.52{\scriptsize\,$\pm$0.29} & 82.86{\scriptsize\,$\pm$0.11} & 52.19{\scriptsize\,$\pm$0.32} & 36.37{\scriptsize\,$\pm$0.43} \\
\bottomrule
\end{tabular}%
}
\end{table}

\end{document}